\documentclass[a4paper,fleqn]{cas-dc}
\renewcommand{\printorcid}{}

\usepackage[numbers]{natbib}
\usepackage{bbding}
\usepackage{balance}

\usepackage{atbegshi}% http://ctan.org/pkg/atbegshi
\AtBeginDocument{\AtBeginShipoutNext{\AtBeginShipoutDiscard}}
\begin{document}
\let\WriteBookmarks\relax
\def\floatpagepagefraction{1}
\def\textpagefraction{.001}
\shorttitle{Object-Based Augmentation Improves Quality of Remote Sensing Semantic Segmentation}
\shortauthors{Svetlana Illarionova et~al.}

\title [mode = title]{Object-Based Augmentation Improves Quality of Remote Sensing Semantic Segmentation}                      
% \tnotemark[1,2]

% \tnotetext[1]{This document is the results of the research
%   project funded by the National Science Foundation.}

% \tnotetext[2]{The second title footnote which is a longer text matter
%   to fill through the whole text width and overflow into
%   another line in the footnotes area of the first page.}

\hyphenation{me-tho-do-lo-gy}

\author[1]{Svetlana~Illarionova}
\cormark[1]
\ead{S.Illarionova@skoltech.ru}.   %!!!!!!!!!!!!!!!!!!
% \ead[url]{www.cvr.cc, cvr@sayahna.org}

% \credit{Conceptualization of this study, Methodology, Software}

\author[1]{Sergey~Nesteruk}

\author[1]{Dmitrii~Shadrin}

\author[1]{Vladimir~Ignatiev}

\author[1]{Mariia~Pukalchik}

\author[1]{Ivan~Oseledets}

\address[1]{Skolkovo Institute of Science and Technology, Moscow 143025, Russia}

% \author[2,4]{Han Theh Thanh}[style=chinese]

% \author[2,3]{CV Rajagopal}[%
%   role=Co-ordinator,
%   suffix=Jr,
% \ead{cvr3@sayahna.org}
% \ead[URL]{www.sayahna.org}

% \credit{Data curation, Writing - Original draft preparation}

% % \address[2]{Sayahna Foundation, Jagathy, Trivandrum 695014, India}

% \author%
% [1,3]
% {Rishi T.}
% \cormark[2]
% \fnmark[1,3]
% \ead{rishi@stmdocs.in}
% % \ead[URL]{www.stmdocs.in}

% \address[3]{STM Document Engineering Pvt Ltd., Mepukada,
%     Malayinkil, Trivandrum 695571, India}

\cortext[cor1]{Corresponding author}
% \cortext[cor2]{Principal corresponding author}
% \fntext[fn1]{This is the first author footnote. but is common to third
% \fntext[fn2]{Another author footnote, this is a very long footnote and
%   it should be a really long footnote. But this footnote is not yet
%   sufficiently long enough to make two lines of footnote text.}

% \nonumnote{This note has no numbers. In this work we demonstrate $a_b$
%   the formation Y\_1 of a new type of polariton on the interface
%   between a cuprous oxide slab and a polystyrene micro-sphere placed

\begin{abstract}
Today deep convolutional neural networks~(CNNs) push the limits for most computer vision problems, define trends, and set state-of-the-art results. In remote sensing tasks such as object detection and semantic segmentation, CNNs reach the SotA performance. However, for precise performance, CNNs require much high-quality training data. Rare objects and the variability of environmental conditions strongly affect prediction stability and accuracy. To overcome these data restrictions, it is common to consider various approaches including data augmentation techniques. This study focuses on the development and testing of object-based augmentation. The practical usefulness of the developed augmentation technique is shown in the remote sensing domain, being one of the most demanded in effective augmentation techniques. We propose a novel pipeline for georeferenced image augmentation that enables a significant increase in the number of training samples. The presented pipeline is called object-based augmentation~(OBA) and exploits objects' segmentation masks to produce new realistic training scenes using target objects and various label-free backgrounds. We test the approach on the buildings segmentation dataset with six different CNN architectures and show that the proposed method benefits for all the tested models. We also show that further augmentation strategy optimization can improve the results. The proposed method leads to the meaningful improvement of U-Net model predictions from $0.78$ to $0.83$ F1-score. 

\end{abstract}

% % \begin{graphicalabstract}
% % \includegraphics{figs/grabs.pdf}
% % \end{graphicalabstract}

% \begin{highlights}
% \item Segmentation masks allow augmenting each object on the image individually.
% \item Object-based image augmentation notably improves model performance on sparse targets and small datasets.
% \item It is possible to improve model performance further by optimizing augmentation hyperparameters.
% \item Enlarging the set of background images increases model generalization to the new regions and conditions.
% \end{highlights}


\maketitle
\setcounter{page}{1}

\section{Introduction}

% Conclusion

Machine learning models depend drastically on the data quality and its amount. In many cases, using more data allows the model to reveal hidden patterns deeper and achieve better prediction accuracy~\cite{Sun_2017_ICCV}. However, gathering of a high-quality labeled dataset is a time-consuming and expensive process~\cite{data2019automating}. %Therefore, limitation on training samples rises in many studies adding extra demands to the neural-network model and the solution approach in general. 
Moreover, it is not always possible to obtain additional data: in many tasks, unique or rare objects are considered~\cite{Nesteruk2021Image_Compression} or access to the objects is restricted~\cite{HUANG2019372}. In other tasks, we should gather data rapidly~\cite{shadrin-2020}. The following tasks are among such challenges: operational damage assessment in emergency situations~\cite{novikov2018satellite}, medical image classification~\cite{masquelin2021wavelet}. There are different approaches to address dataset limitations: pseudo labeling, special architectures development, transfer learning~\cite{zhang2015initialize},~\cite{bullock2019xnet},~\cite{barz2020deep},~\cite{ng2015deep}. 
Another standard method to address this issue is image augmentation. Augmentation means applying transformations (such as flip, rotate, scale, change brightness and contrast) to the original images to increase the size of the dataset~\cite{info11020125}. 

In this study, we focus on augmentation techniques for the remote sensing domain. Lack of labeled data for particular remote sensing tasks makes it crucial to generate more training samples artificially and prevent overfitting~\cite{8113128}. However, simple color and geometrical transformations fail to meet all geospatial demands. The goal of this work is to propose an object-based augmentation~(OBA) pipeline for the semantic segmentation task that works with high-resolution georeferenced satellite images. Naming our augmentation methodology object-based, we imply that this technique targets separate objects instead of whole images. The idea behind the approach is to crop objects from original images using their masks and pasting them to a new background. Every object and background can be augmented independently to increase the variability of training images; shadows for pasted objects also can be added artificially. We show that our approach is superior to the classic image-based methods in the remote sensing domain despite its simplicity. The pipeline is tested in a building segmentation task using U-Net~\cite{ronneberger2015u} and Feature Pyramid Network~(FPN)~\cite{lin2017feature} with three different encoder sizes to reveal a relationship between convolutional neural network~(CNN) architecture and augmentation benefit. 

The main contributions of this paper are:
\begin{itemize}
    \item We propose a novel for remote sensing domain simple and efficient augmentation scheme called OBA that improves CNN model generalization for satellite images.
    \item We test the proposed method on the building segmentation task and show that our approach outperforms common augmentation approaches.
    \item We show that OBA parameters can be efficiently optimized for better performance.
\end{itemize}

The underlying code with experiments will be shared.

The remaining paper is organized as follows: Section~\ref{sec:related} describes common augmentation approaches for remote sensing problems; Section~\ref{method}~describes object-based augmentation methodology and the practical way to tune augmentation hyperparameters; Section~\ref{experiment}~illustrates the experiments scheme; Section~\ref{result}~reports the results and the influence of the training pipeline, and dataset size on the final score.

\section{Related works}
\label{sec:related}

% tasks for ml
%Augmentation is often implemented in tasks of semantic segmentation, classification, and object detection. Although there are specificities in each of the tasks, the widespread augmentation strategy is geometrical and color transformations. 

% State of the art
We can split all image augmentations into two groups according to the target. Image-based augmentations transform the entire image. On the contrast, object-based augmentation technique targets every object in the image independently~\cite{ghiasi2020simple}, \cite{nesteruk2021image}. It makes augmentations more flexible and provides a better way to handle sparse objects which is particularly useful for remote sensing problems. However, this novel approach has not been studied yet in the remote sensing domain.

 %In~\cite{}, to address the classification problem, they clipped images belonging to the same class and created new ones from such crops. 
In~\cite{zoph2020learning}, for the object detection task, they perform image transformations individually within and outside bounding boxes. They also change bounding box position regarding the background. In~\cite{zhou2019slot} authors clip area with the target object and replace it with the same class object from another image. However, the bounding box's background is still from the source image of the new object. It makes generated image less realistic and can affect further classification. In~\cite{ghiasi2020simple}, for semantic segmentation, authors use objects' masks to create new images with pasted objects.  %Other important augmentation approaches are described in~\cite{wang2019perspective}, .

% Mention a notable alternative approach
Another notable augmentation approach is based on generative adversarial neural networks~(GANs)~\cite{ostankovich2020application}. It generates completely new training examples that can benefit the final model performance~\cite{frid2018synthetic}. However, this approach requires training an auxiliary model that produces training samples for the main model. In this work, we focus only on augmentation approaches that require neither major changes in the training loop nor much computational overhead.

% Areas where augmentation is used
Augmentation is extensively used in various areas.  In~\cite{perez2018data}, they proposed augmentation for medical images aimed to classify skin lesions. In~\cite{xu2017underwater}, augmentation was implemented for underwater photo classification. Another sphere of study that processes images distinguished from regular camera photos is remote sensing~\cite{HUANG2019215}.

% Remote sensing
%In the training process, flip and rotation were adopted to augment the training set. \cite{HUANG2019215}
The most frequently used augmentation approach in remote sensing is also color and geometrical transformations~\cite{yu2017deep}, \cite{zong2019deep}, \cite{korez2020weighted}, \cite{illarionova2020neural}. In~\cite{rs11040403}, rescaling, slicing, and rotation augmentations were applied in order to increase the quantity and diversity of training samples in building semantic segmentation task. In~\cite{ren2018small}, authors implemented "random rotation" augmentation method for small objects detection. In~\cite{stivaktakis2019deep}, authors discussed advances of augmentation leveraging in the landcover classification problem with limited training samples. Another task and augmentation approach is described in~\cite{yan2019data}. The authors used 3D ship models to insert them into the background obtained from high-resolution satellite images. Another augmentation approach with 3D models leveraging for aircraft detection was described in~\cite{yan2019novel}. The main limitation of listed works is related to 3D models' unavailability for most remote sensing problems. The above overview clearly states the importance of the augmentation techniques in current computer vision research as well as high capabilities for making the trained models more generalized and precise. Thus, the improvement of the augmentation techniques is crucial for the development of accurate solutions for practical computer vision tasks.

% The conducted literature review shows that image augmentation in the remote sensing domain has to be studied further. % ... (more reasons)      

%Code is available through the link: \href{https://github.com/LanaLana/satellite_object_augmentation}{\color{blue}{https://github.com/LanaLana/satellite\_object\_augmentation}}.

\iffalse
\begin{figure*}[htbp]
    \centering
    \includegraphics[width=0.5\columnwidth]{orig_pilot.png}\includegraphics[width=0.5\columnwidth]{orig_pilot.png}\includegraphics[width=0.5\columnwidth]{orig_pilot.png}
    \caption{Background images examples.}
    \label{fig:background}
\end{figure*}
\fi

\section {METHODOLOGY}
\label{method}

\begin{figure*}[htbp]
    \centering
    \includegraphics[width=1.4\columnwidth]{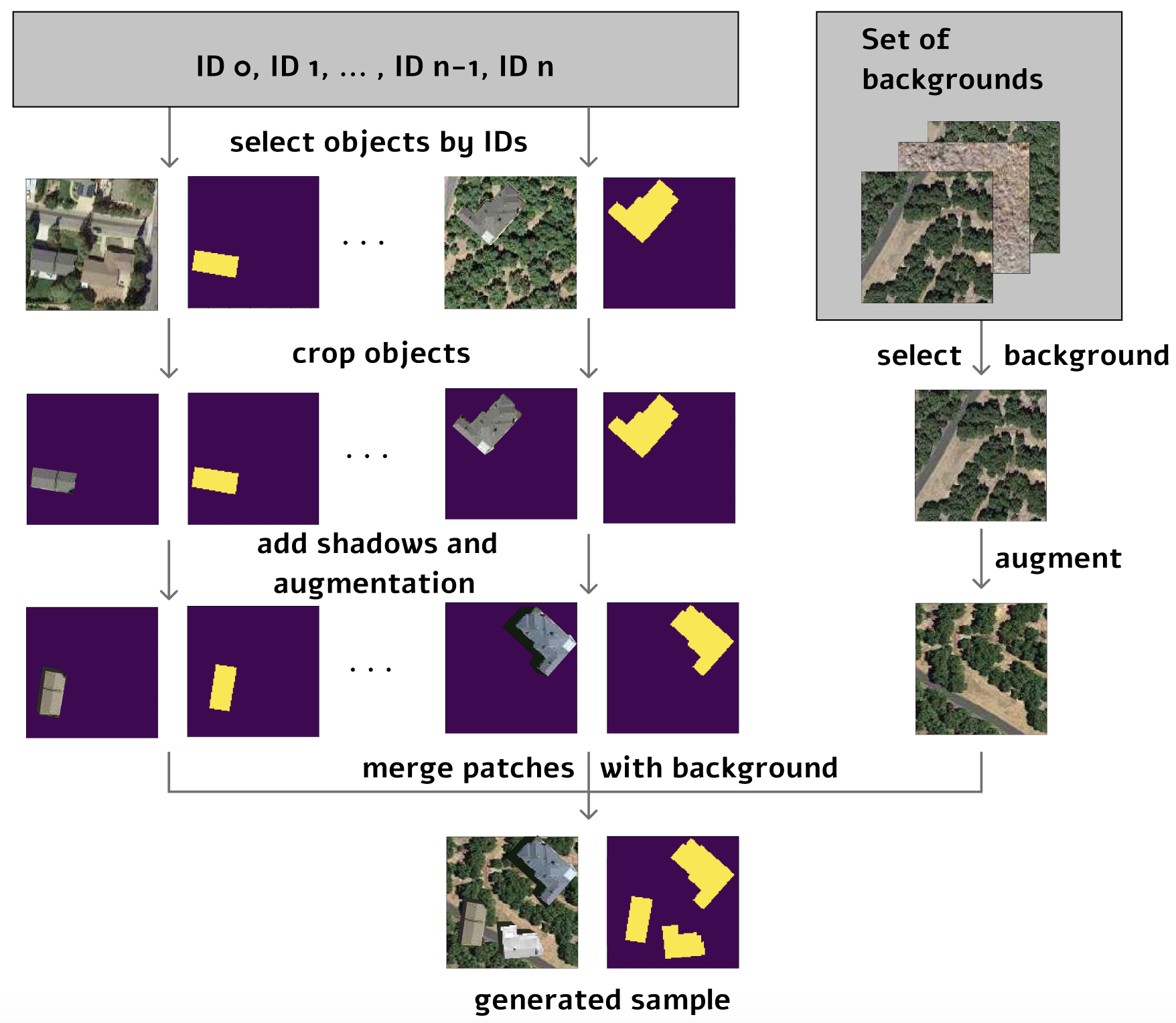}
    \caption{Object-based augmentation (OBA) scheme. For each generated sample, we choose objects from the set of IDs, crop objects according to its footprint, add shadows, conduct geometrical and color transformations, and then merge these cut objects with a new background.}
    \label{fig:workflow}
\end{figure*}

\begin{figure*}[htbp]
    \centering
    \includegraphics[width=1.8\columnwidth]{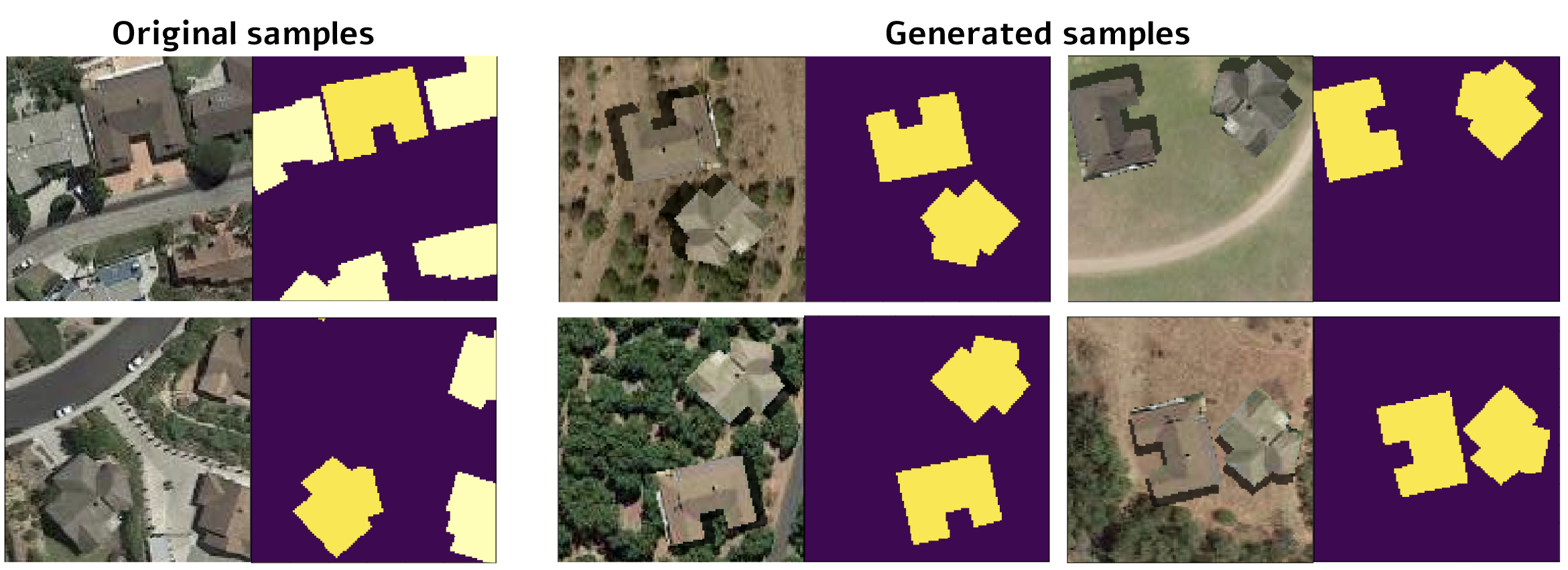} %building_full
    \caption{Examples of augmented samples reconstructing various environmental conditions. Objects and backgrounds are from different images and have different color and geometrical transformations. Shadows are added artificially for the generated samples. }
    \label{fig:example_aug}
\end{figure*}

% \begin{figure}[h!]
%     \centering
%     \includegraphics[width=0.7\columnwidth]{building1.png}
    
%     \includegraphics[width=0.7\columnwidth]{building2.png}
    
%     \includegraphics[width=0.7\columnwidth]{building3.png}
%     \caption{Examples of augmented samples (buildings and background are from different images).}
%     \label{fig:example_aug}
% \end{figure}

\subsection{Object-based augmentation}

This section describes the object-based augmentation methodology for the semantic segmentation problem. 

Object-based augmentation requires images containing objects with masks and background images. Each object has its ID and shape coordinates extracted from a geojson file. There were two types of background areas: from the initial dataset and new unlabeled images that aim to add diversity into data. An object and a background were chosen randomly. According to the object's coordinates, a crop with a predefined size containing the object was clipped for RGB channels and masks~(Figure~\ref{fig:workflow}). The background crop has the same size. With some set probability, the object's crop and the background's crop were augmented separately or together using base color and geometrical transformations from Albumentations package~\cite{info11020125}. This package is popular both for semantic segmentation task in general and remote sensing domains. The considered in our study transformations are described in Table~\ref{tab:album_transforms}. Since most of the works in remote sensing do not specify albumentations parameters for images augmentation~\cite{zong2019deep}, \cite{korez2020weighted}, we also set default parameters.

\begin{table}[htbp]
\centering
\caption{BASE COLOR AND GEOMETRICAL TRANSFORMATIONS FROM ALBUMENTATIONS PACKAGE.}
    \label{tab:album_transforms}
\begin{tabular}{ll}

\textbf{Transformation} & \textbf{Description} \\

RandomRotate90 & Randomly rotate the input by 90 \\ & degrees zero or more times \\
Flip & Flip the input either horizontally,  \\ & vertically or both  \\
Additive & Add Gaussian noise to \\ GaussianNoise &  the input image \\
HueSaturationValue & Randomly change hue, \\ &  saturation and value of the \\ &  input image \\
CLAHE & Apply Contrast Limited Adaptive \\ &  Histogram Equalization to the \\ &  input image \\
OpticalDistortion & Apply Barrel Distortion~\cite{gribbon2003real} \\ &  to the image \\
RandomContrast & Randomly change contrast of  \\ & the input image \\
RandomBrightness & Randomly change brightness \\ &  of the input image \\
IAAEmboss & Emboss the input image and overlays \\ &  the result with the original image \\
MotionBlur & Apply motion blur to the input \\ &  image using a random-sized kernel \\

\end{tabular}
\end{table}

% \textit{RandomRotate90, Flip, IAAAdditiveGaussianNoise, HueSaturationValue, CLAHE, OpticalDistortion, RandomContrast, RandomBrightness, RandomGamma, IAAEmboss, MotionBlur}. 

The object extension was then merged with a new background by placing it in a random position strictly within the image crop. Objects number within each image crop was chosen randomly in a predefined range. Overlapping between objects was prohibited. 

To make generated samples more realistic, we add shadows using objects' footprints~(Figure~\ref{fig:example_aug}). The mask of the shadowed area is blended with initial background pixels with different intensities.

The entire new sample generating process is conducted during model training. It aimed to ensure greater diversity without memory restrictions related to additional sample storage. Therefore, all functions for object-based augmentation were implemented into the data-loader and generator. New generated samples are also alternated with original samples.

In summary, OBA includes the following options:

\begin{itemize}
    \item Shadows addition (length and intensity may vary);
    \item Objects number per crop selection (default: up to 3 extra objects); %  (default: up to 3 extra objects);
    \item Selection of base color and geometrical transformations probability (default: $50\%$); %  (default: $70\%$);
    \item Background images selection (default: $60\%$); %  (default: $60\%$);
    \item Selection of original and generated samples mixing probability (default: $60\%$).  %  (default: $60\%$). 
\end{itemize}

We compared this augmentation approach with the following alternatives: 
\textit{Baseline} is training a CNN model using just base color and geometrical augmentations~(from Albumentations framework);
\textit{Baseline\_no\_augm} is training a CNN model without any augmentations;
\textit{OBA\_no\_augm} is applying OBA cropping and pasting without generic augmentations;
\textit{OBA\_no\_shadow} is applying OBA cropping and pasting without adding generated shadows to the objects; 
\textit{OBA\_no\_background} is applying OBA cropping and pasting using a background from the same image only.
Note that all the tested OBA approaches except A use crops from other images to form a background. The summary of the experiments is reflected in Table~\ref{tab:exp}.

\begin{table*}[htbp]
\centering
\caption{Augmentation approaches comparison for different training set size using U-Net with Resnet34 encoder (F1-score  for the test set).}
    \label{tab:results_buildings}
\begin{tabular}{l|lll|lll|llr}
\hline
 & \multicolumn{3}{c}{\textit{Baseline\_no\_augm} } & \multicolumn{3}{c}{\textit{Baseline}}  & \multicolumn{3}{c}{\textit{OBA}} \\
  % & \multicolumn{3}{c}{Without}  & \multicolumn{3}{c}{Base}  &  \multicolumn{3}{c}{Object-based augm +}  \\
 %  & \multicolumn{3}{c}{  augmentation} &  \multicolumn{3}{c}{augmentation}  &   \multicolumn{3}{c}{extra background}   \\
\hline 
Training set size   & 1/3  & 2/3 & 1   & 1/3  & 2/3 & 1  &  1/3  & 2/3 & 1 \\
\hline 
Original test set       & 0.415  & 0.43  & 0.45    & 0.751  & 0.785  & 0.788  & 0.787  & 0.81  & 0.829   \\
Generated    test set   & 0.508  & 0.58  & 0.79 &   0.696 &  0.747 &  0.907  & 0.941  & 0.954  & 0.957  \\ 

\hline
\end{tabular}
\end{table*}

\begin{table*}[htbp]
\centering
\caption{Augmentation approaches comparison for different CNN models (F1-score for the test set).}
    \label{tab:nn_size_buildings_full}
\begin{tabular}{l|l|lll|lll|llr}
\hline
  \multirow{2}{*}{Model} & & \multicolumn{3}{c}{\textit{Baseline\_no\_augm}}  & \multicolumn{3}{c}{\textit{Baseline}}  & \multicolumn{3}{c}{\textit{OBA}} \\
 %  &  & \multicolumn{3}{c}{  augmentation} &  \multicolumn{3}{c}{augmentation}  &  \multicolumn{3}{c}{extra background}   \\
 \cline{2-11}
 & Backbone  & Resnet18  & Resnet34 & Resnet50  & Resnet18 & Resnet34 & Resnet50 & Resnet18  & Resnet34 & Resnet50  \\
\hline 

\multirow{4}{*}{FPN}  &  Original       & \multirow{2}{*}{0.325}   & \multirow{2}{*}{0.367}   & \multirow{2}{*}{0.186}  & \multirow{2}{*}{0.741}  & \multirow{2}{*}{0.762}  & \multirow{2}{*}{0.784}  & \multirow{2}{*}{0.802}  & \multirow{2}{*}{0.813} & \multirow{2}{*}{\textbf{0.826}}   \\
&   test       &    &   &   &   &   &   &   &   &    \\

&  Generated       & \multirow{2}{*}{0.418}   & \multirow{2}{*}{0.388}   & \multirow{2}{*}{0.255}  & \multirow{2}{*}{0.641}  & \multirow{2}{*}{0.606}  & \multirow{2}{*}{0.585}  & \multirow{2}{*}{0.935}  & \multirow{2}{*}{0.95} & \multirow{2}{*}{\textbf{0.958}}   \\
&   test       &    &   &   &   &   &   &   &   &    \\

\hline

\multirow{4}{*}{U-Net}  &  Original       & \multirow{2}{*}{0.435}   & \multirow{2}{*}{0.45}   & \multirow{2}{*}{0.34}  & \multirow{2}{*}{0.766}  & \multirow{2}{*}{0.788}  & \multirow{2}{*}{0.766}  & \multirow{2}{*}{0.807}  & \multirow{2}{*}{\textbf{0.829}} & \multirow{2}{*}{0.824}   \\
&   test       &    &   &   &   &   &   &   &   &    \\

&  Generated       & \multirow{2}{*}{0.422}   & \multirow{2}{*}{0.49}   & \multirow{2}{*}{0.40}  & \multirow{2}{*}{0.744}  & \multirow{2}{*}{0.907}  & \multirow{2}{*}{0.585}  & \multirow{2}{*}{0.932}  & \multirow{2}{*}{\textbf{0.947}} & \multirow{2}{*}{\textbf{0.947}}   \\ % 49 79
&   test       &    &   &   &   &   &   &   &   &    \\

\hline

\end{tabular}
\end{table*}
\subsection{Optimization}

% Policy
The task of optimal augmentation policy choice is a significant part of algorithm adjustment. Many works are devoted to this topic~\cite{fawzi2016adaptive},~\cite{lim2019fast},~\cite{cubuk2020randaugment}. It is often defined as a combinatorial optimization problem of optimal transformations search within some available set.

To choose the best augmentation strategy, we set experiments with the optimizer from the Optuna software framework~\cite{optuna_2019}. It uses a multivariant Tree-structured Parzen Estimator. Optuna helps to search hyperparameters efficiently and shows significant improvements for various machine learning and deep learning tasks~\cite{haddad20203d}, \cite{kato2021predicting}. The optimizer supports earlier pruning to reject weak parameters initialization. As a pruner, we used Optuna's implementation of MedianPruner. Loss function value after each epoch was evaluated to choose new parameters' values in the searching space. For object-based augmentation, the following parameters were considered: 
\begin{itemize}
    \item Number of generated objects within one crop; 
    \item Probability of the base color transformations;
    \item Probability of object-based augmentation;
    \item Probability of extra background usage.
\end{itemize}

For this study, we set $12$ epochs and the same validation samples representation without any modifications to obtain the most equivalent criteria as possible for earlier pruning.

%\begin{figure}[h!]
%    \centering
%    \includegraphics[scale=.75]{plot.png}
%    \caption{Dependence between dataset size and model accuracy}
%    \label{fig:plot}
%\end{figure}

\iffalse
\begin{figure*}[h!]
    \centering
    \includegraphics[width=0.8\columnwidth]{gen_ch_0.jpg}
    \includegraphics[width=0.8\columnwidth]{gen_ch_1.jpg}
    \caption{Generated and original NIR band (GAN)}
    \label{fig:nir_channel}
\end{figure*}
\fi

\section{EXPERIMENTS}
\label{experiment}  % \label{} allows reference to this section

\subsection{Dataset}

\begin{figure*}[htbp]
    \centering
    \includegraphics[width=1.\columnwidth]{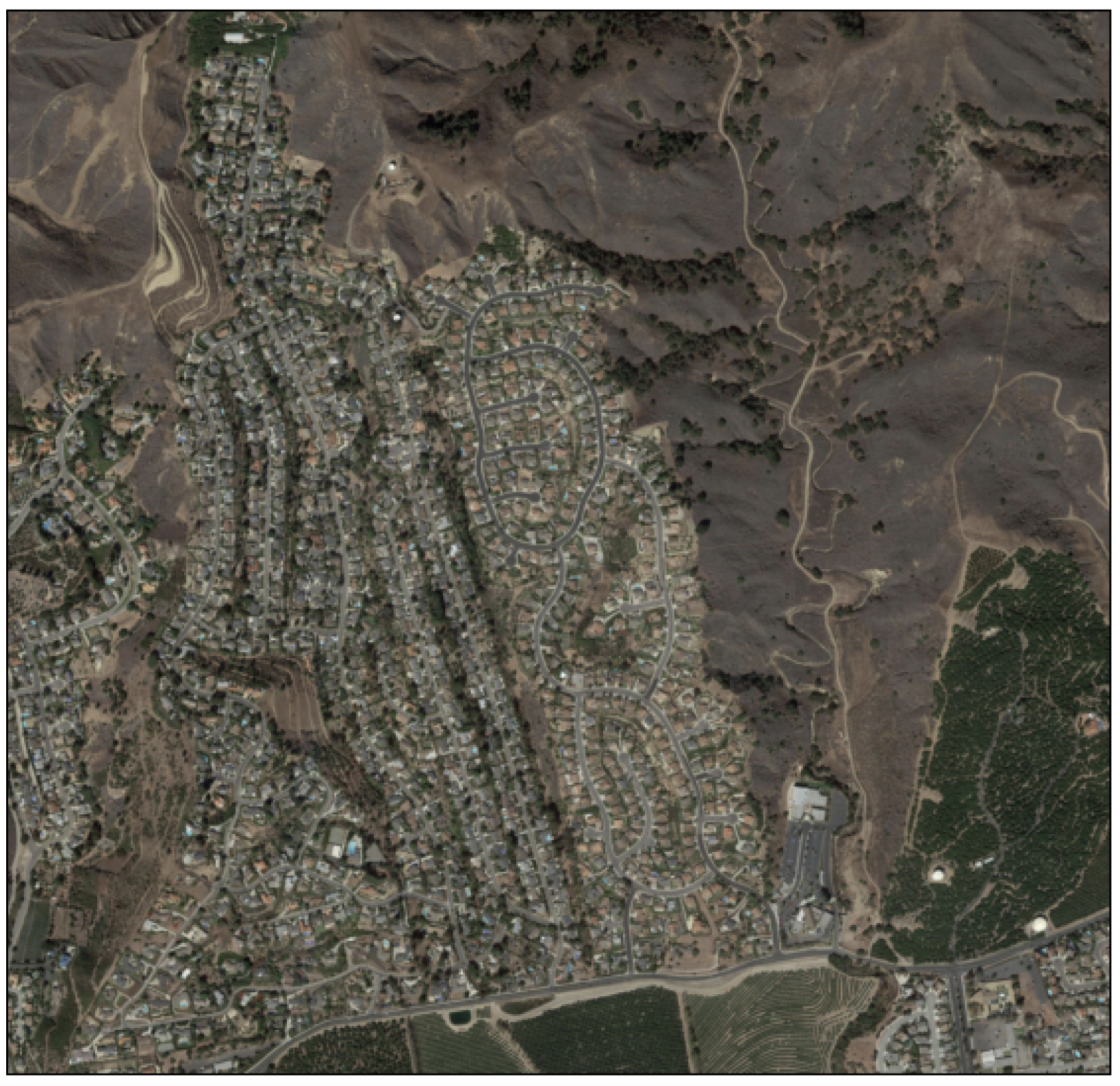} \includegraphics[width=1.\columnwidth]{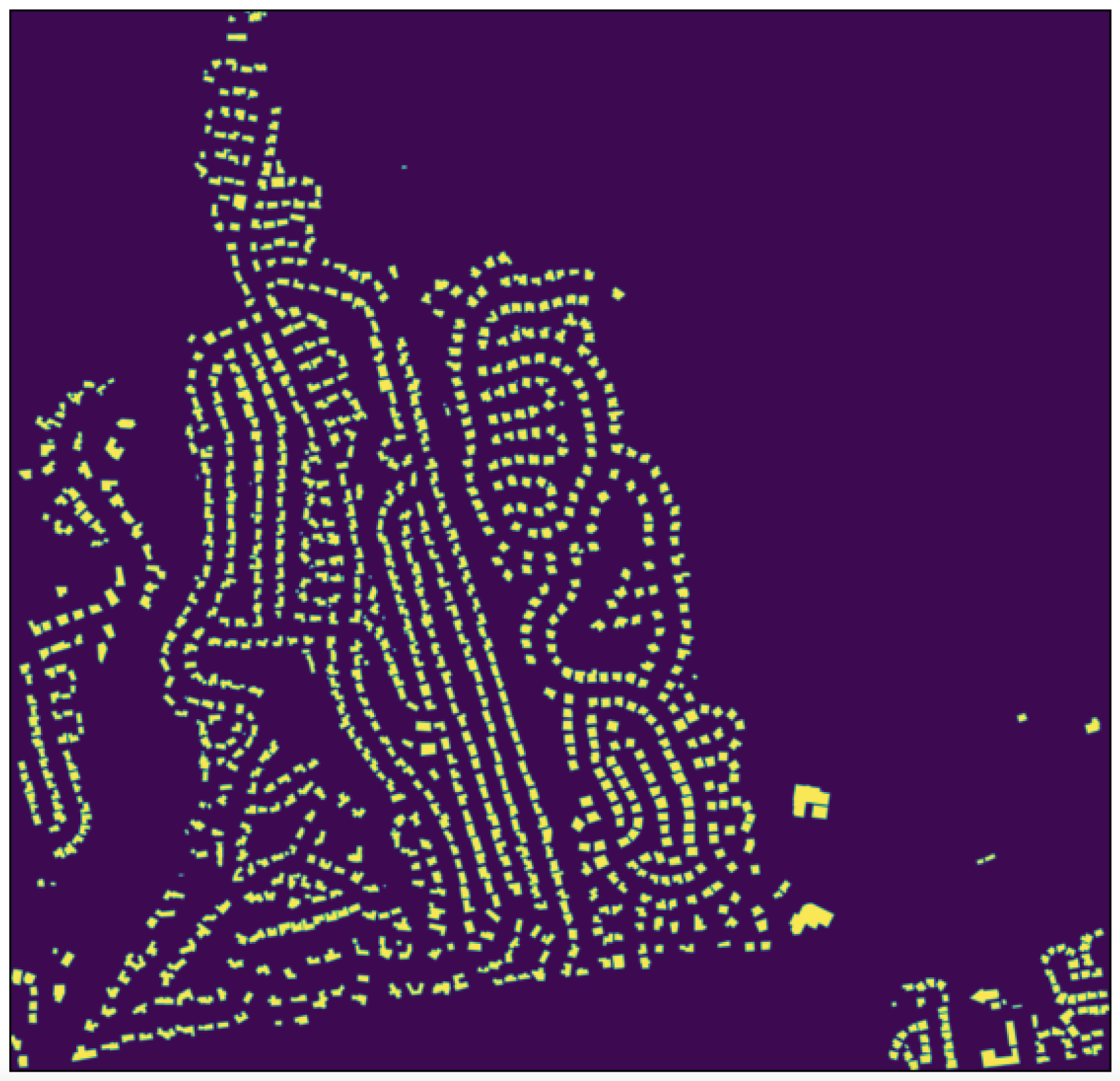}
    \caption{Train area and mask. The image size is $4418*4573$ pixels}
    \label{fig:train_area}
\end{figure*}

\begin{table}[htbp]
\caption{Dataset description.}
    \label{tab:split}
\begin{center}
\begin{tabular}{l|c|c|r}
\hline
  & Train & Validation  & Test  \\

\hline
Objects number   & 955  & 226 & 282  \\
Area in hectars & 390 & 100 & 93 \\
Extra background area & 2000 & 500 & 500 \\ % ...
in hectars &  &  &  \\
\hline
\end{tabular}
\end{center}
\end{table}

\begin{table}[htbp]
\caption{Experiments with different augmentation setups.}
    \label{tab:exp}
\begin{center}
\begin{tabular}{l|c|c|c}
\hline
  & Base   & Shadow &  Extra  \\
 & augm.   &  &  background  \\
\hline
\textit{Baseline\_no\_augm }             & {\color{Maroon} \XSolidBrush }  & {\color{Maroon} \XSolidBrush } & {\color{Maroon} \XSolidBrush } \\
\textit{Baseline }      &  {\color{ForestGreen} \Checkmark }  & {\color{Maroon} \XSolidBrush } & {\color{Maroon} \XSolidBrush } \\
\textit{OBA\_no\_augm }       & {\color{Maroon} \XSolidBrush }  & {\color{ForestGreen} \Checkmark }  & {\color{ForestGreen} \Checkmark } \\
\textit{OBA\_no\_shadow }    & {\color{ForestGreen} \Checkmark }   & {\color{Maroon} \XSolidBrush } & {\color{ForestGreen} \Checkmark } \\
\textit{OBA\_no\_background } & {\color{ForestGreen} \Checkmark }   & {\color{ForestGreen} \Checkmark }   & {\color{Maroon} \XSolidBrush } \\
\textit{OBA }              &  {\color{ForestGreen} \Checkmark } & {\color{ForestGreen} \Checkmark }  & {\color{ForestGreen} \Checkmark } \\
\hline
\end{tabular}
\end{center}
\end{table}

We evaluated the developed augmentation pipeline in the remote sensing semantic segmentation problem, namely the buildings segmentation task. It is an important problem for remote sensing, and it was considered in different studies~\cite{8710471}, \cite{rs11040403}. Lack of labelled training data makes it suitable for the OBA approach evaluation.

 For building segmentation, we used the dataset described in~\cite{novikov2018satellite}. This dataset was collected for damage assessment in the emergency and included images before and after wildfires in California in 2017. However, we leveraged just data before the event. It covers Ventura and Santa Rosa counties (the total area is about $580$ hectares). Very high-resolution RGB images for this region were available through Digitalglobe within their Open Data Program. We used $955$ buildings for training and $226$ for validation from Ventura and $282$ buildings from Santa-Rosa for the test (see Table \ref{tab:split}). Objects' masks are presented both in raster TIFF format and vector shapes. Training image is shown in Figure \ref{fig:train_area}. 
 
 %For large vehicles segmentation, we used a subset of DOTA dataset~\cite{Xia_2018_CVPR}. It consists of very high spatial resolution satellite images supplemented with masks of target objects. Each image is georeferenced. Objects' masks are represented as vector shapes. Dataset includes ... images and ... objects.  

 We selected high-resolution extra background without target objects from Maxar serves \cite{maxar} (image id is \textit{lnu-lightning-complex-fire}, April 15, 2020, California). We cut test, validation and train images with the total area of about $3000$ ha. It includes various land-cover types such as: lawns, individual trees, roads, and forested areas.

\subsection {Effect of the train dataset size}

To assess the effect of the dataset size on the final model score we considered the following samples portions for training dataset:

\begin{itemize}
    \item The entire training dataset;
    \item 2/3 of the entire training dataset;
    \item 1/3 of the entire training dataset.
\end{itemize}

For each experiment, we fixed the same validation set that was not reduced further. For the reduced training dataset, we ran a model on different subsets. There were 2 and 3 subsets for each of the mentioned dataset sizes. The final results for each training subset size were defined as an average. We conducted these experiments for three different training modes: without augmentation (\textit{Baseline\_no\_augm}), with base color and geometrical transformations (\textit{Baseline}), with object-based augmentation (\textit{OBA}).    

% \subsection{Augmented and original dataset size proportion}

To evaluate how original and generated samples affect the final score the following experiment was conducted:

\begin{enumerate}
    \item Pretrain model using just generated samples for predefined fixed number of epochs: 5, 10, 15, or 20;
    \item Continue training using just original samples for predefined fixed number of epochs: 2, 4, or 8.
\end{enumerate}

Therefore, we aimed to obtain 12 models that utilize for training different proportions of generated and original samples. Such methodological experiments allow us to obtain results that provide important information on the best possible training strategy in order to achieve the highest score. Also, results are useful for performing further analysis of the sensitivity of models performance and training procedure to the developed augmentation technique which in turn allow using the most beneficial aspects of the proposed augmentation algorithm.

% \begin{table}[h!]
% \caption{Augmentation pretraining. F1-score for initial test dataset. 5, 10, 15, 20 -- epochs with augmentation, 2, 4, 8 -- final epochs without augmentation.}
%     \label{tab:strategy_initial}
% \begin{center}
% \begin{tabular}{l|c|c|r}
% \hline
% Epochs  & \textit{2} & \textit{4} & \textit{8}  \\

% \hline
% \textit{5}   & 0.742 & 0.774 & 0.727 \\
% \textit{10} & 0.638 & \textbf{0.79} & 0.739 \\
% \textit{15}   & 0.708 & 0.736 & 0.747 \\
% \textit{20} & 0.72 & 0.747 & 0.763 \\
% \hline
% \end{tabular}
% \end{center}
% \end{table}

% \begin{table}[h!]
% \caption{Augmentation pretraining. F1-score for augmented test dataset. 5, 10, 15, 20 -- epochs with augmentation, 2, 4, 8 -- final epochs without augmentation.}
%     \label{tab:strategy_background}
% \begin{center}
% \begin{tabular}{l|c|c|r}
% \hline
%  Epochs & \textit{2} & \textit{4} & \textit{8}  \\

% \hline
% \textit{5}   & 0.802 & 0.839 & 0.758     \\
% \textit{10} & 0.79 & \textbf{0.842} & 0.738 \\
% \textit{15}   & 0.798 & 0.807 & 0.67 \\
% \textit{20} & 0.795 & 0.805 & 0.745 \\
% \hline
% \end{tabular}
% \end{center}
% \end{table}

\begin{table}[htbp]
\caption{F1-score results with augmentation pretraining, and fine-tuning on original data (U-Net with ResNet-34 encoder).}
    \label{tab:strategy_full}
\begin{center}
\begin{tabular}{l|c|c|c|c|c|r}
\hline

% & \multicolumn{3}{c|}{Results on initial set} & \multicolumn{3}{c}{Results on augmented set} \\

& \multicolumn{3}{c|}{Results on} & \multicolumn{3}{c}{Results on} \\

& \multicolumn{3}{c|}{initial set} & \multicolumn{3}{c}{augmented set} \\

\cline{2-7}
 & \multicolumn{6}{c}{Fine-tuning epochs} \\

\hline
Pretrain   & \multirow{2}{*}{2} & \multirow{2}{*}{4} & \multirow{2}{*}{8}  & \multirow{2}{*}{2} & \multirow{2}{*}{4} & \multirow{2}{*}{8} \\
epochs & & & & & & \cr

\hline
\textit{5}   & 0.742 & 0.774 & 0.727        & 0.802 & 0.839 & 0.758 \\
\textit{10} & 0.698 & \textbf{0.795} & 0.739     & 0.79 & \textbf{0.842} & 0.738 \\
\textit{15}   & 0.708 & 0.736 & 0.747     & 0.798 & 0.807 & 0.67 \\
\textit{20} & 0.72 & 0.747 & 0.763     & 0.795 & 0.805 & 0.745  \\
\hline
\end{tabular}
\end{center}
\end{table}

%\subsection{Training}

\subsection{Neural Networks Models And Training Details}

To evaluate the object-based augmentation approach on different fully  convolutional neural networks  architectures, we considered FPN~\cite{lin2017feature} and U-Net~\cite{ronneberger2015u} with three encoders' sizes: ResNet-18, ResNet-34, ResNet-50~\cite{he2016deep}. Both U-Net and FPN are popular CNN architectures for semantic segmentation tasks in remote sensing domain~\cite{KATTENBORN202124}, \cite{LI2020296}. All models used weights pre-trained on “ImageNet” classification dataset~\cite{deng2009imagenet}. We used models’ architecture implementation from~\cite{Yakubovskiy:2019}.

The training of all the neural network models was performed at a PC with GTX-1080Ti GPUs. For each model, the following training parameters were set. An RMSprop optimizer with a learning rate of $0.001$, which was reduced with the patience of $3$. There were $50$ (except the experiment with augmentation strategies) epochs with $100$ steps per epoch and $30$ steps for validation. Early stopping was chosen with the patience of $4$, then the best model according to validation score was considered. The batch size was specified to be $30$ with a crop size of $128*128$ pixels. Such a crop size is a typical choice in remote sensing tasks with CNN models~\cite{KATTENBORN202124}, \cite{liu2020detection}. The batch size was chosen according to GPU memory limitations. As a loss function, binary cross entropy (Equation~\ref{eq:ce_loss}) was used.  

\begin{equation}
    \begin{aligned}
        % L(X_i, Y_i) = - \sum_{j=1}^{c} y_{ij} * log(p_{ij}),
        L(y, \hat{y}) = - \frac{1}{N} \sum_{i=1}^{N} y_i log(\hat{y_i}) + (1 - y_i) log(1 - \hat{y_i}), 
    \end{aligned}
    \label{eq:ce_loss}
\end{equation}

where: \\
$ N $ is the number of target mask pixels; \\ 
$ y $ is the target mask; \\ 
$ \hat{y} $ is the model prediction. \\

\subsection{Evaluation}

The model outputs were binary masks of target objects, which were evaluated against the ground truth with pixel-wise F1-score (Equation~\ref{eq:f1}). F1-score is robust for unsymmetrical datasets, and it is a commonly used score for semantic segmentation tasks~\cite{csurka2013good}, in particular in the remote sensing domain~\cite{KATTENBORN202124}. 

\iffalse
\begin{equation}
    \begin{aligned}
        F1 = \frac{2*precision*recall}{precision + recall},
    \end{aligned}
    \label{eq:f1}
\end{equation}

\begin{equation}
    \begin{aligned}
        precision = \frac{TP}{TP+FP}, \\
    \end{aligned}
    \label{eq:precision}
\end{equation}

\begin{equation}
    \begin{aligned}
        recall = \frac{TP}{TP+FN} \\
    \end{aligned}
    \label{eq:recall}
\end{equation}
\fi

\begin{equation}
    \begin{aligned}
        % precision = \frac{TP}{TP+FP}, recall = \frac{TP}{TP+FN}, \\
        % F1 = \frac{2*precision*recall}{precision + recall}
        F_1 = \frac{TP}{TP + \frac{1}{2} (FP + FN)},
    \end{aligned}
    \label{eq:f1}
\end{equation}

where $TP$ is True Positive (number of correctly classified pixels of the given class), $FP$ is False Positive (number of pixels classified as the given class while in fact being of other class, and $FN$ is False Negative (number of pixels of the given class, missed by the method).
% \fi

To evaluate model performance, two types of reference images were used. The first one was the initial image from the dataset. The second one was the artificially generated test image that includes objects from the original test image and a new background image without target objects. The background image was split into the grid with cell size $128*128$ pixels. A random object with generated shadows of different sizes and intensities was pasted to each cell with the probability $60\%$. The position within a cell was also chosen randomly.    

For each experiment, we run a CNN model three times with different random seeds and averaged results.

\section{RESULTS AND DISCUSSION}
\label{result}

\begin{figure}[htbp]
    \centering
    \includegraphics[width=1.\columnwidth]{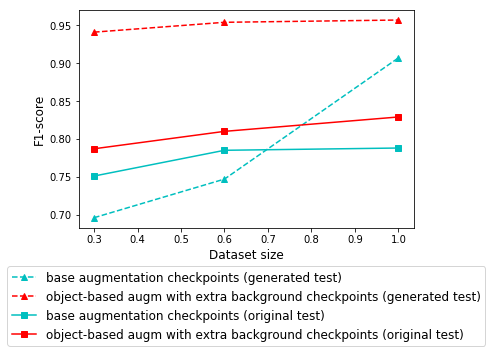} 
    \caption{Relationship between F1-score on test image and training dataset size, U-Net with ResNet-34 encoder.}
    \label{fig:dataset_size}
\end{figure}

\begin{figure*}[htbp]
     \centering
         \centering
         
        \includegraphics[width=2.0\columnwidth]{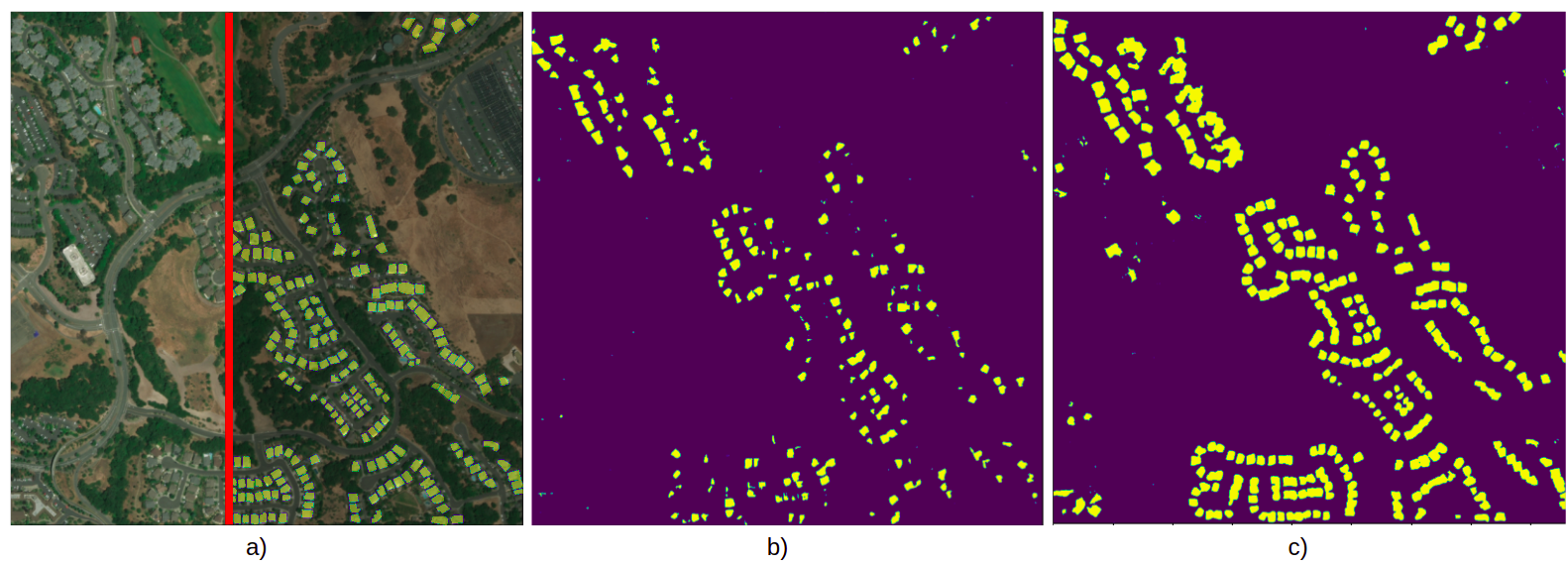}

        \caption{Sample results on the test set of buildings dataset: a) input RGB with ground truth on the right of the red line; b) prediction without augmentation; c) prediction with object-based augmentation.}
        \label{fig:pred_final_wv}
\end{figure*}

\begin{table}[htbp]
\caption{Experiments with different augmentation setups (F1-score for the test set, U-Net with ResNet-34 encoder).}
    \label{tab:diff_features}
\begin{center}
\begin{tabular}{c|l|c}
\hline
Standard  & \multirow{2}{*}{Augmentation}  &  \multirow{2}{*}{F1-score}\\
augmentation  &  & \\
\hline
\multirow{2}{*}{No} & \textit{Baseline\_no\_augm} & 0.45 \\
 & \textit{OBA\_no\_augm} & 0.66 {\color{blue}(+21\%)} \\
\hline
\multirow{5}{*}{Yes} & \textit{Baseline } & 0.788  \\
 &\textit{OBA\_no\_shadow } & 0.811 (+2.3\%) \\
 & \textit{OBA\_no\_background } & 0.81 (+2.2\%)  \\
 & \textit{OBA } & 0.829 (+4.1\%)  \\
 & \textit{OBA + optimization} & 0.835 {\color{blue}(+4.7\%)} \\
\hline
\end{tabular}
\end{center}
\end{table}

% \begin{figure}[!htbp]
%      \centering
%      \begin{subfigure}[b]{0.4\textwidth}
%          \centering
%          \includegraphics[scale=.3,width=\columnwidth]{ buildings_rgb.png}
%          %\caption{Input image.}
%      \end{subfigure}
%      \hfill
%      \begin{subfigure}[b]{0.4\textwidth}
%          \centering
%          \includegraphics[scale=.35,width=\columnwidth]{ gt_santa.png}
%          %\caption{Ground truth.}
%      \end{subfigure}
%      \hfill
%      \begin{subfigure}[b]{0.4\textwidth}
%          \centering
%          \includegraphics[scale=.35,width=\columnwidth]{ pred_santa.png}
%          %\caption{Prediction with base augmentation.}
%      \end{subfigure}
%      \hfill
%      \begin{subfigure}[b]{0.4\textwidth}
%          \centering
%          \includegraphics[scale=.35,width=\columnwidth]{ base_pred.png}
%          %\caption{Prediction with object-based augmentation.}
%      \end{subfigure}
%         \caption{Example results on the test set of buildings dataset. Input RGB, ground truth, prediction with object-based augmentation, prediction without augmentation}
%         \label{fig:pred_final_wv}
% \end{figure}

% \begin{figure}[!htbp]
%      \centering
%          \centering
         
%         \includegraphics[width=0.8\columnwidth]{santa_fulll_vert.png}

%         \caption{Example results on the test set of buildings dataset. Input RGB, ground truth, prediction with object-based augmentation, prediction without augmentation}
%         \label{fig:pred_final_wv}
% \end{figure}

\subsection{Object-based augmentation}

We compared different augmentation approaches and presented results in Table~\ref{tab:results_buildings}. Model predictions for the test region are presented in Figure~\ref{fig:pred_final_wv}. \textit{OBA} allows us to improve the F1-score for the entire dataset size from~$0.788$ to~$0.829$ for the original test set and from~$0.907$ to~$0.935$ for generated test set compared with the base color and geometrical transformations (\textit{Baseline}). As experiments clearly indicate, the model trained without any data augmentation (\textit{Baseline\_no\_augm}) performed significantly poorly~($0.45$~F1-score). 

Extra background usage improves prediction quality compared with models that use only initial background areas both for the original test set (F1-score from~$0.81$ to~$0.829$) and generated test set (F1-score from~$0.907$ to~$0.957$). Additional backgrounds make a model more universal for new regions. It is promising in cases where we want to switch between different environmental conditions without extra labeled datasets. 

Even without extra background images, remote sensing task specificity frequently offers an opportunity to add more diversity in training samples. Target objects are often too small compared with the entire satellite image that is leveraged for a particular task. Moreover, target objects can be distributed not regularly which creates large areas free of them. We show that even these areas can be successfully used to create new various training samples (see \newline \textit{OBA\_no\_background} in Table \ref{tab:diff_features}). 

We studied artificial shadows importance in the proposed approach. As shown in Table \ref{tab:diff_features}), shadows allow us to improve the model performance from $0.811$ to $0.829$ (F1-score). Therefore, a shadow is an essential descriptor for objects observed remotely from satellites. It distinguishes OBA for remote sensing tasks from the copy-paste approach~\cite{ghiasi2020simple} applied in the general computer vision domain.  

We tested the proposed approach with different neural networks architectures. The results of the experiments are shown in Table~\ref{tab:nn_size_buildings_full}. Models with different capacities perform better on different tasks; however, for both U-Net and FPN architectures with different encoders sizes, the object-based approach outperforms the base augmentation strategy on the original test set. Generated test set helps to estimate model generalization to new environmental conditions. Object-based augmentation clearly improves generalization in our experiments. %It is also worth noticing that models trained without any augmentation have very poor performance (F1-score $0.45$).

% sampling

Results for different proportions of original and generated samples are presented in Table~\ref{tab:strategy_full}. It indicates that pretraining in the object-based augmentation mode~(without original sample usage) for~$10$ epochs and further training in the base augmentation mode for~$4$ epochs leads to the best result for the considered experiment. This experiment indicates that separate training on the original and generated data during different epochs is not an optimal choice in this task. More efficient approach is to set a probability to add augmented and original samples into each batch during training. 

The advantage of this method is that it does not require much computational overhead. It needs just one model training on the generated dataset and tuning the model from several checkpoints on the original dataset. However, as Table~\ref{tab:nn_size_buildings_full} shows, the strategy of mixing generated and original images during the training process leads to better results than separating image sources.

As we evaluated the object-based augmentation approach for remote sensing tasks with man-made objects, one of the future study directions is to implement the described method to wider classes, in particular, vegetation objects, such as agricultural crops or individual trees.

\subsection{Optimization}

For the optimization task, we tested U-Net with ResNet-34 encoder using the entire dataset. Optuna package was leveraged to find better values for augmentation parameters, namely, extra objects number ($[0, 1, 2, 3]$), the probability to use additional background ($0-1$), object-based augmentation probability ($0-1$), and color augmentation probability ($0-1$). We run $20$ trials; for each trial, parameter values varied. For the optimization process, Optuna utilized loss function values on the validation set after each epoch. As the pruner method, we used MedianPruner. 

Augmentation strategy search for U-Net model increases the final performance from $0.829$ to $0.835$ (\textit{OBA} and \textit{OBA + optimization} in \ref{tab:diff_features}) for the original test and from $0.947$ to $0.955$ for the generated test. The found optimal parameters are as follows: 

\begin{itemize}
    \item \textit{extra objects} $= 3$;
    \item \textit{background prob} $= 0.53$;
    \item \textit{object-based augmentation probability} $= 0.787$;
    \item\textit{color augmentation probability} $= 0.35$.
\end{itemize}

As it is shown, the optimizer allows us to set up better augmentation parameters according to the particular task specificity. 

\subsection{Dataset size}

The results for experiments with different dataset sizes are present in Figure~\ref{fig:dataset_size}. Object-based augmentation allows avoiding the drastic drop in prediction quality when dataset size is reduced. For buildings segmentation with object-based augmentation, dataset size decreasing to one-third leads to F1-score decreasing from $0.957$ to $0.941$, while with the base augmentation it decreases from $0.907$ to $0.696$. That makes object-based augmentation suitable for few-shot learning, especially when high-capacity models are used. 

% ~\cite{8710471}, \cite{rs11040403}

As a possible application, the proposed method can be used for models training on other buildings datasets to study larger training set size and new environmental conditions. For instance, the SpaceNet building dataset~\cite{van2018spacenet} considered in~\cite{8710471}, \cite{rs11040403}. 

%Table~\ref{tab:nn_size_buildings_full}

\section{Conclusion}
\label{sect:conclusion}

This study proposes an advanced object-based augmentation approach that outperforms standard color and geometrical image transformations. The presented method combines target objects from georeferenced satellite images with new backgrounds to produce more diverse realistic training samples.  
We also explicate the importance of augmentation hyperparameters tuning and describe a practical way to find optimal object-based augmentation parameters.
Our results show promising potential for real-life remote sensing tasks making CNN models more robust for new environmental conditions even if the labeled dataset size is highly limited.

% \section*{Acknowledgment}
% The authors acknowledge the use of the Skoltech CDISE supercomputer Zhores for obtaining the results presented in this paper.

%While basic data augmentation has contributed to CNN prediction quality in various remote sensing tasks, this study proposed a more advanced object-based augmentation approach that outperforms standard color and geometrical image modifications. The presented method combines target objects from georeferenced satellite images and new backgrounds to produce more diverse realistic training samples. 
%It shows promising potential for real-life remote sensing tasks making CNN models more robust for new environmental conditions even if the labeled dataset size is highly limited. 

%\acknowledgments 
%This unnumbered section is used to identify those who have aided the authors in understanding or accomplishing the work presented and to acknowledge sources of funding.  

%%%%% References %%%%%

% \bibliographystyle{IEEE}
% \bibliography{report}

\bibliographystyle{elsarticle-harv} 
\bibliography{report.bib}


\bio{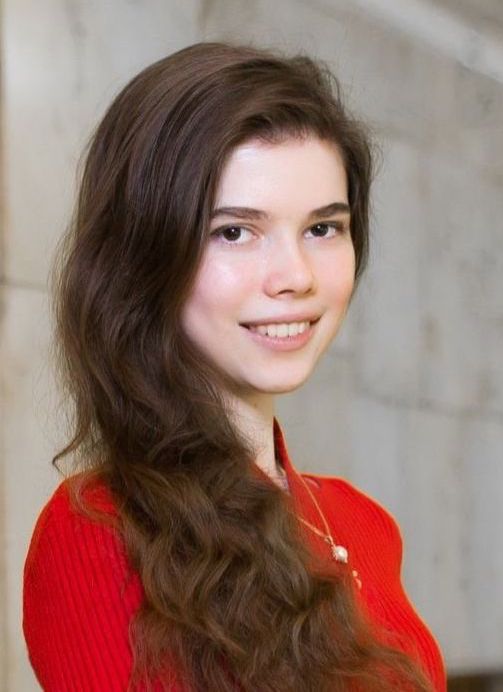}
Svetlana Illarionova received the Bachelor and Master degrees in computer science from Lomonosov Moscow State University, Moscow, Russia, in 2017 and 2019, respectively. She is currently working toward the Ph.D. degree in computer science at Skolkovo Institute of Science and Technology, Moscow, Russia. Her research interests include computer vision, deep neural networks, and remote sensing.
\endbio

\newpage

\bio{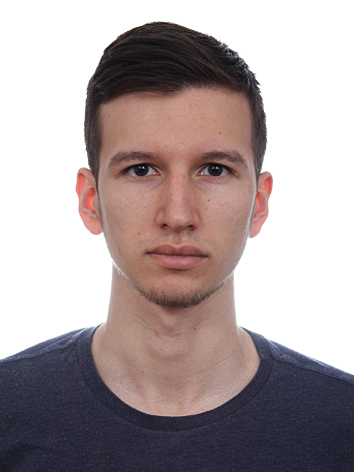}
Sergey Nesteruk is a PhD student at the Skolkovo Institute of Science and Technology (Skoltech), Russia. Sergey received his BS and MS in Information Security at Saint Petersburg University of Aerospace Instrumentation in 2018 and 2020, respectively. In 2020 he also received his MS degree in Information Science and Technology at Skoltech. Sergey's research is related to monitoring systems and applying Machine Learning methods to the collected data. Sergey is involved in the development of Precision Agriculture Lab at Skoltech and is responsible for the development of greenhouse image collecting systems, development of image augmentation framework, and computer vision research. 
\endbio

\bio{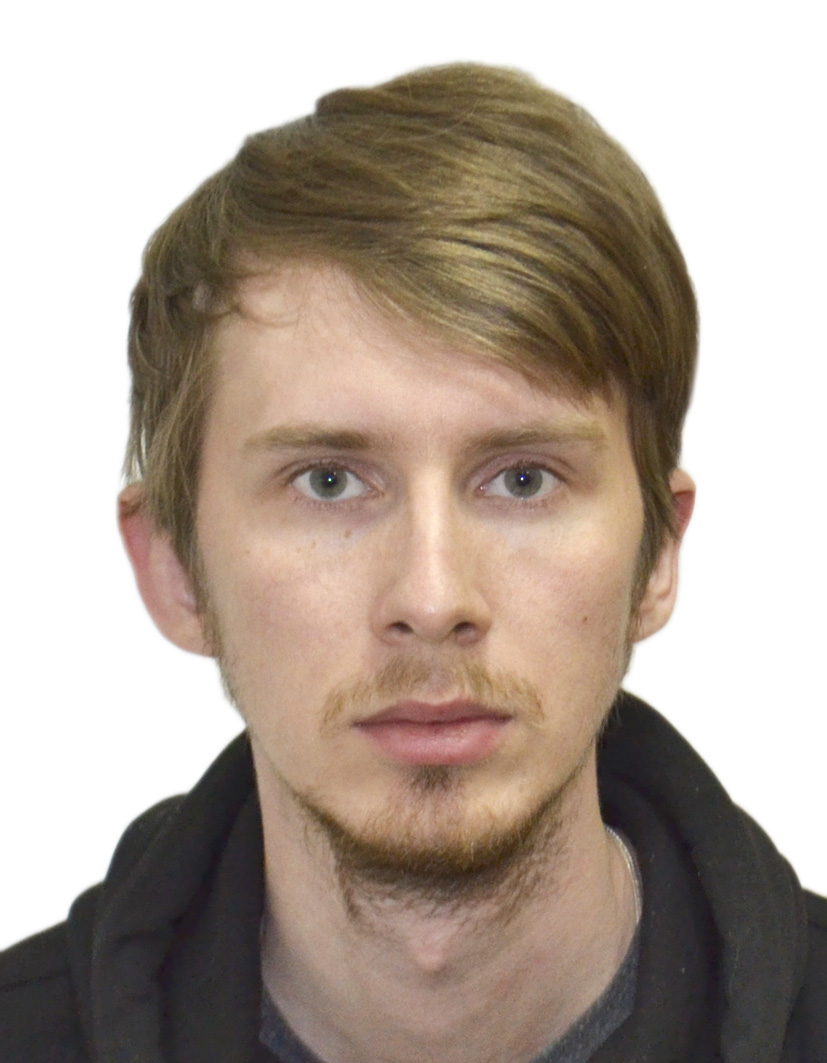}
Dmitrii Shadrin is a Researcher at the Skolkovo Institute of Science and Technology (Skoltech), Russia. Dmitrii received his PhD in Data Science from Skoltech in 2020 and MS in Applied Physics and Mathematics at the Moscow Institute of Physics and Technology  (MIPT) in 2016. His research interests include data processing, modelling of physical and bio processes in closed artificial growing systems, machine learning, and computer vision. Dmitry is involved in the development of Digital Agriculture Lab at Skoltech and is responsible for the experimental research and a number of projects in the lab.
\endbio

%\newpage

\bio{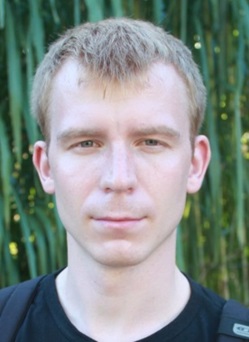}
Vladimir Ignatiev is a research scientist at Skolkovo Institute of Science and Technology, Moscow, Russia. He leads the Aeronet Lab that is focused on various applications of the deep learning methods to remote sensing data. Vladimir graduated from Moscow Institute of Physics and Technology in 2012, and defended his Ph.D. thesis in 2017. Before joining Skoltech, he worked in Dorodnitsyn CCAS and Aerocosmos Research Institute. He has experience in different remote sensing data processing and forecasting models development.
\endbio

\bio{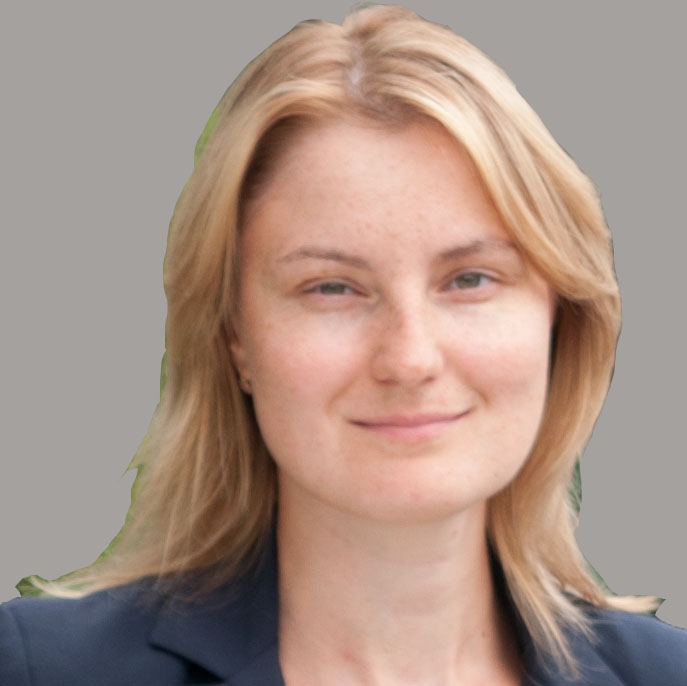}
Mariia Pukalchik received the M.Sc. degree in ananlytical chemistry from the Vyatka State University, Russia, in 2009, and the Ph.D. degree in Ecology from the Lomonosov Moscow State University (MSU), Russia, in 2013. In 2017, she joined the Skoltech, where she is currently an Assistant Professor. She has authored 20 refereed articles and two books. Maria has experience in the field of of Machine Learning and Artificial intelligence application in biomedical, environmental, and agriculture fields. Mariia is involved in the development of Digital Agriculture Lab at Skoltech. 
\endbio

\bio{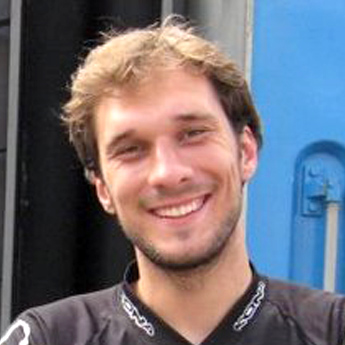}
Ivan Oseledets graduated from Moscow Institute of Physics and Technology in 2006, got Candidate of Sciences degree in 2007, and Doctor of Sciences in 2012, both from Marchuk Institute of Numerical Mathematics of Russian Academy of Sciences. He joined Skoltech CDISE in 2013. 
Ivan’s research covers a broad range of topics. He proposed a new decomposition of high-dimensional arrays (tensors) – tensor-train decomposition, and developed many efficient algorithms for solving high-dimensional problems. His current research focuses on development of new algorithms in machine learning and artificial intelligence such as construction of adversarial examples, theory of generative adversarial networks and compression of neural networks. It resulted in publications in top computer science conferences such as ICML, NIPS, ICLR, CVPR, RecSys, ACL and ICDM.
Professor Oseledets is an Associate Editor of SIAM Journal on Mathematics in Data Science, SIAM Journal on Scientific Computing, Advances in Computational Mathematics (Springer).
\endbio

\balance

\end{document}